\documentclass[letterpaper, 10 pt, conference]{ieeeconf}  

\IEEEoverridecommandlockouts                              

\usepackage{graphicx}

\newtheorem{definition}{Definition}
\usepackage{amsmath}
\usepackage{amsfonts}
\usepackage{amssymb}
\usepackage{algorithm}
\usepackage{algpseudocode}

\title{\LARGE \bf
On the Optimality, Stability, and Feasibility of Control Barrier Functions: An Adaptive Learning-Based Approach
}

\author{Alaa Eddine Chriat$^{1}$ and Chuangchuang Sun$^{1}$
\thanks{$^{1}$The authors are with the Aerospace Engineering Department, Mississippi State University, Starkville, MS 39759, USA. Emails:
        {\tt\small aec652@msstate.edu, csun@ae.msstate.edu}.}%
}

\usepackage{xcolor}

\graphicspath{{figures/}}

\usepackage{amsfonts}
\usepackage{microtype}
\usepackage{graphicx}
\usepackage{subfigure}
\usepackage{booktabs} 
\usepackage{balance}
\usepackage{makecell}
\usepackage{bm}
\makeatletter
\let\NAT@parse\undefined
\makeatother
\usepackage{hyperref}
\usepackage[numbers,sort&compress]{natbib}

\let\labelindent\relax

\usepackage{amssymb}
\usepackage{multirow}
\usepackage{amsmath}
\usepackage{algorithm,algpseudocode}
\usepackage{wrapfig}

\newcommand{\bea}{\begin{eqnarray}}
\newcommand{\eea}{\end{eqnarray}}
\newcommand{\beas}{\begin{eqnarray*}}
\newcommand{\eeas}{\end{eqnarray*}}
\newcommand{\leftm}{\left[\begin{array}}
\newcommand{\rightm}{\end{array}\right]}

\newcommand{\mA}{\mathcal{A}}

\usepackage{diagbox,hhline}
\usepackage{enumitem}

\def\span{{\mbox{\bf span}}}

\newcommand{\mS}{\mathcal{S}}

\newcommand{\mE}{\mathbb{E}}

\definecolor{commentclr}{RGB}{110, 149, 204}

\usepackage{stackengine}
\stackMath

\usepackage{color}
\usepackage{soul}
\usepackage{xspace}
\usepackage{cleveref}
\usepackage{makecell}
\usepackage{balance}

\begin{document}

\maketitle
\thispagestyle{empty}
\pagestyle{empty}

\begin{abstract}
Safety has been a critical issue for the deployment of learning-based approaches in real-world applications. To address this issue, control barrier function (CBF) and its variants have attracted extensive attention for safety-critical control. However, due to the myopic one-step nature of CBF and the lack of principled methods to design the class-$\mathcal{K}$ functions, there are still fundamental limitations of current CBFs: optimality, stability, and feasibility. 
In this paper, we proposed a novel and unified approach to address these limitations with Adaptive Multi-step Control Barrier Function (AM-CBF), where we parameterize the class-$\mathcal{K}$ function by a neural network and train it together with the reinforcement learning policy. Moreover, to mitigate the myopic nature, we propose a novel \textit{multi-step training and single-step execution} paradigm to make CBF farsighted while the execution remains solving a single-step convex quadratic program. Our method is evaluated on the first and second-order systems in various scenarios, where our approach outperforms the conventional CBF both qualitatively and quantitatively.

\end{abstract}

\section{INTRODUCTION}\label{sec:intro}
While (deep) learning-based approaches have been pervasive nowadays, safety issues limit their deployment in real-world applications, especially those with humans in the loop.
For example, autonomous driving vehicles should guarantee the safety of the drivers and other entities by following the driving rules. Other safety-critical applications can be found in industrial, medical, and household scenarios.
Therefore, learning-enable models should rigorously guarantee safety, and failing to do so can result in undesirable or even disastrous outcomes.

\begin{figure}[t]
  \centering
  \includegraphics[scale=.475]{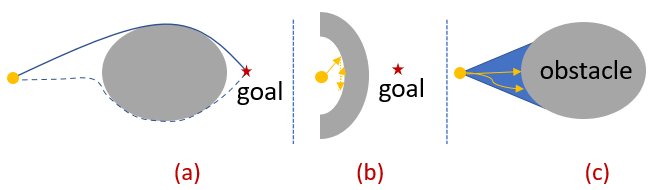}
  \caption{Limitations of CBF. (a) One step is often myopic and thus generates an overall sub-optimal path.
           (b) When marching towards the goal driven by the control Lyapunov functions, the CBF agent gets stuck into the trap. (c) Limited translational/angular control input fails to avoid the obstacle for high-order systems.}
  \label{cbf limitations}
  \vspace{-.75cm}
\end{figure}

In recent years, the control barrier function (CBF~\cite{ames2016control}) has attracted extensive attention due to its forward invariance property and scalability of solving a convex quadratic programming (QP) such that many variants have been developed in different settings and application scenarios.
Additionally, the combination of reinforcement learning (RL) and control barrier functions \cite{ames2016control, cheng2019end, zheng2020safe, choi2020reinforcement, alshiekh2017safe,fulton2018safe, turchetta2020safe, garcia2015comprehensive} attracts much attention for safety assurance and explorations by using CBF as the safety shield. 
Specifically, work in \cite{safeoffpolicyRL} integrates the CBF into the utility function of RL,
while others have used neural networks to parameterize and learn the barrier function parameters~\cite{neuralbarriercertificate,xiao2023barriernet}. Moreover, some other works integrated model predictive control with CBF as a predictive safety filter for reinforcement learning\cite{enhancedsafetymechanism}. However, while control barrier functions are widely investigated and studied, there are still major issues addressed as follows. (i) The one-step forward nature, while rendering simplicity and scalability, also makes it myopic. (ii) The goal-reaching and safety guarantee, driven by control Lyapunov functions (CLF) and CBF, can often conflict with each other. (iii) The barrier function $\kappa(\bullet)$ is often manually designed (such as linear and quadratic candidates) and thus lacks expressiveness and adaptivity. 
Furthermore, such issues lead to the following concrete limitations; see the illustrations in Fig.~\ref{cbf limitations}. (1) It can often lead to an overall sub-optimal controller design, with ``greedy'' single-step control synthesis. 
(2) Because of the one-step planning nature to minimize the control Lyapunov function, it can easily get trapped in a concave safety set. For example, when an autonomous vehicle tries to go through an intersection of two convex obstacles, it can get stuck there due to the objective to minimize the CLF. (3) It can often encounter infeasibility \cite{xiao2022sufficient} due to control limitations in high-order systems. In other words, the CBF constraint conflicts with the control constraints. A common example is the adaptive cruise control scenario when it is "too late to brake" when deceleration is limited such that collision cannot be avoided. We aim to address those fundamental challenges in CBF via a learning-based approach.

Learning and control approaches have been combined closely to mitigate their respective disadvantages while keeping the advantages. Modern control theory has rigorous guarantees of stability and constraint satisfaction with accurate dynamics models given. Such guarantees are often missing in partial-observable environments, with pervasive noise and uncertainty. Moreover, the design of proper metrics, such as Lyapunov functions,
is often case-by-case and requires expert knowledge. A principled way to design such metrics is desirable. As a result, data-driven learning-based control has attracted much attention in recent years. Methods are developed to learn the unmodelled dynamics and quantify the uncertainty, such as the Gaussian process \cite{rodriguez2021learning, khan2021safety, peng2021trajectory}. Lyapunov function \cite{chow2018lyapunov, berkenkamp2017safe, richards2018lyapunov} and (neural) contraction metric \cite{tsukamoto2021learning, tsukamoto2020neural, tsukamoto2021learning2, tsukamoto2021theoretical} based methods are developed to guarantee the stability of the dynamical systems. As a result, a learning-based adaptive multi-step control barrier function method
is proposed to improve expressiveness, optimality, feasibility, and optimality for the control of safety-critical autonomous systems. Specifically, we propose to learn a class-$\mathcal{K}$ function in a principled way. Moreover, for the myopic nature of CBF, we propose a novel \textit{muli-step training and single-step execution} paradigm. Intuitively, in training it considers a long horizon (instead of one step) and in execution/ inference, the advantage of single-step QP is kept. This, to the best of our knowledge, is the first systematic and unified approach toward those dundamental issues.

\begin{figure*}[thpb]
  \centering
  \includegraphics[scale=.5]{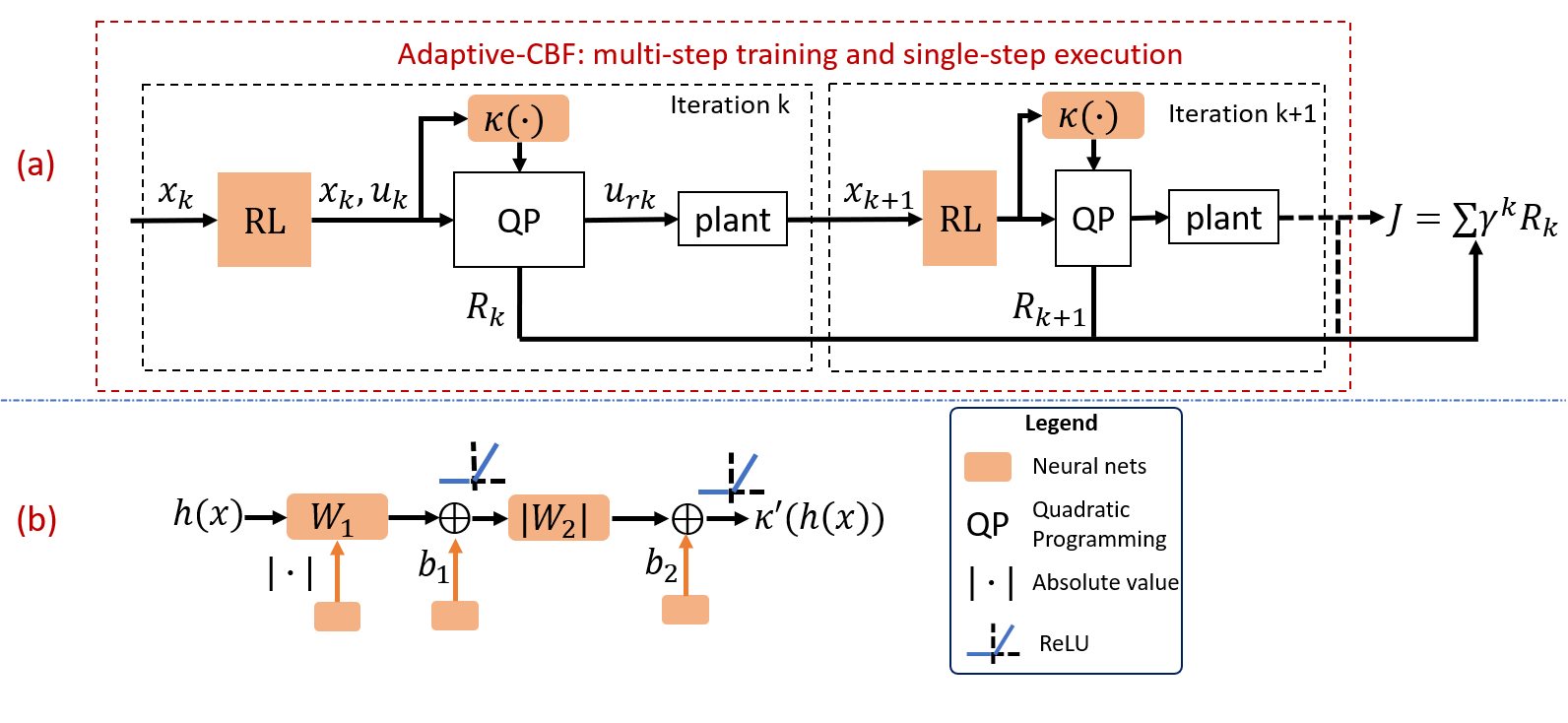}
  \vspace{-0.5cm}
  \caption{Overview of the adaptive multi-step control barrier function (AM-CBF). (a) An end-to-end
trainable multi-step CBF via differentiable programming. Back propagating through all the learnable modules, including the $\kappa(\bullet)$ within the quadratic programming, the return $J(\theta)$ will be maximized. (b) The neural network architecture to learn an adaptive class-$\mathcal{K}$ function.}
  \label{AMCBF map}
\end{figure*}

\section{PRELIMINARIES}
\subsection{High-order CBF}
Control barrier functions are used in control theory to guarantee that a dynamical system can achieve some desired goals while remaining within safe constraints. A CBF is a function that quantifies the system’s safety measurements. Hence, we aim to find an control input that keeps the system within its safe set measured by CBF. Mathematically, consider the nonlinear control-affine system:

\begin{equation}\label{nonlinearsystem}
\dot{x}(t)=f(x(t))+g(x(t)) u(t)
\end{equation}
where $f$ and $g$ are globally Lipschitz, $x\in\mathbb{R}^n$ and $u\in\mathbb{R}^m$ are the states and control inputs, respectively, constrained in closed sets, with initial condition $x(t_0) = x_0$.

\begin{definition}
\cite{ames2016control}$h: \mathbb{R}^{n} \rightarrow \mathbb{R}$ is a barrier function for the set $C=\left\{x \in \mathbb{R}^{n}: h(x) \geqslant 0\right\}$ if $\exists$ an extended class-$\mathcal{K}$ function $\alpha(\bullet)$ such that:
\begin{equation}
\begin{gathered}
\sup _{u \in U}[L_{f}h(x)+L_{g}h(x) u+\alpha(h(x))] \geqslant 0 \\
\inf_{\text{int}(C)}[\alpha(h(x)) ]\geqslant 0 \text {  \quad  and  \quad  } \lim_{\partial C} \alpha(h(x))=0
\end{gathered}
\end{equation}
\end{definition}
Because not all systems are first-order in inputs, we can use higher-order control barrier functions to constrain higher-order systems.
\begin{definition}
\cite{xiaohigh}For the non linear system \eqref{nonlinearsystem} with the $m^{th}$ differentiable function $h(x)$ as a constraint, we define a sequence of functions $\psi_{i}$ with $i \in \{1,2,...,m\}$, starting from $\psi_{0}=h(x)$:
\begin{equation}\label{Hcbfform}
\psi_{i}(x, t)=\dot{\psi}_{i-1}(x, t)+\alpha_{i}\left(\psi_{i-1}(x, t)\right) 
\end{equation}
and define $C_{i}(t)$ sequence of safe sets associated with each $\psi_{i}$:
\begin{equation}
C_{i}(t)=\left\{x \in \mathbb{R}^{n}: \psi_{i-1}(x, t) \geqslant 0\right\}
\end{equation}
the function $h(x)$ is a high order control barrier function if there exist extended class-$\mathcal{K}$ functions $\alpha_{i}(\bullet)$ such that:
\begin{equation}
\psi_{m}(x, t) \geqslant 0 
\end{equation}
\end{definition}

CBFs have great potential in designing safe and robust systems, and they have been applied to various applications such as robotics, and autonomous vehicles.

\subsection{Reinforcement learning}

Reinforcement learning (RL) is to learn a policy for sequential decision-making from active interaction with the dynamic systems~\cite{sutton2018reinforcement}. Such dynamic systems are often defined as Markov decision processes (MDP) that can either be fully or partially observable. 
An MDP is a tuple $\langle \mS,\mA,\mathcal{O}, \mathcal{T}, {R}, \gamma, P_0 \rangle$, where $\mS$ is a set of agent states in the environment, $\mA$ is a set of agent actions, $\mathcal{O}$ is a set of observations in partially observable case, $\mathcal{T}:\mS\times \mA\times \mS\to[0,1]$ is the transition function, ${R}$ is the reward function, $\gamma\in[0,1]$ is the discount factor and $P_0:\mS\to [0,1]$ is the initial state distribution. In the partially observable case, the agent receives an observation $o_i$ correlated with the state $s_i$ as $\mS\mapsto \mathcal{O}$.
A policy $\pi: \mS\mapsto P (\mA)$ is a mapping from the state space to probability over actions. $\pi_\theta(a|s)$ denotes the probability of taking action $a$ under state $s$ following a policy parameterized by $\theta$. 
The objective is to maximize the cumulative reward: 
$
J(\theta) = \mE_{\tau\sim p_\theta(\tau)}[\sum_t \gamma^t {R}(s_t, a_t)],
$
where $\tau$ are the trajectories sampled under $\pi_\theta(a|s)$. In order to optimize the policy that maximizes $J(\theta)$, the policy gradient with respect to $\theta$ can be computed as
$
\nabla_\theta J(\theta) = \mE_{\tau\sim \pi_\theta(\tau)}[\nabla_\theta \log\pi_\theta(\tau) G(\tau)],
$
where $G(\tau) = \sum_{t} \gamma^t {R}(s_t, a_t)$~\cite{sutton2018reinforcement}.
The Q-function of a policy $\pi$ is defined as $Q^{\pi}:\mS\times \mA\to {R}$ at any state action pair $(s,a)$. Mathematically, for a policy $\pi$, $Q^{\pi}(s_0, a_0) = \mE_{\pi}[\sum_{t=0}^{\infty} \gamma^t {R}(s_t, a_t)]$ denotes the expected return of the trajectory.
The policy can be deterministic in the form as $\mu_\theta: \mS \mapsto \mA$. As the objective gradient depends on the differentiation over actions, it requires continuous action space. With the policy parameters $\theta$ as deep neural networks (DNN), it is termed as \textit{deep deterministic policy gradient} (DDPG) and can be used as a suitable instantiation of the RL algorithm for continuous control.

\subsection{Differentiable convex programming}
Differentiable convex programming is a technique that allows computation of the gradients of an optimization problem objective function with respect to the parameters of the problem, by taking matrix differentiation of the Karush-Kuhn-Tucker (KKT) conditions. One example of a differentiable optimization method is OPTNET~\cite{amos2017optnet}, which has differentiable optimization problems within the architecture of the neural network. During training, the gradients of the objective function are back-propagated through the neural network. In general, we can use this method to differentiate through any disciplined convex program~\cite{diffcvxoptlayer}, by mapping it into a cone program first~\cite{diffconeprog}, computing the gradients, and mapping back to the original problem. A common example of differentiable programming is learning the constraints of the optimization problem such as convex polytopes or ellipsoid projections, through supervised learning.
The advantage of differentiable optimization methods like OPTNET is that they can be used to optimize a wide range of convex objectives that are difficult to optimize using traditional optimization methods.

\section{APPROACH: ADAPTIVE MULTI-STEP CONTROL BARRIER FUNCTION}

\subsection{Learning-based CBF: a multi-step training and single-step execution paradigm}
Control barrier functions have been used to enforce safety constraints in control systems. In reinforcement learning, CBFs can be used to ensure that an agent's actions satisfy safety constraints while maximizing a reward function. In general, CBFs can be used as a safety shield that projects an unsafe action into a safe one via the CBF conditions~\cite{zheng2020safe, cheng2019end, choi2020reinforcement}. However, non-learning-based CBFs suffer the limitations described in \Cref{sec:intro}, which we aim to address here with its learning-based counterpart.


Consider the nonlinear system \eqref{nonlinearsystem}, the objective of safe reinforcement learning is to generate a policy/control $u_r$ to achieve certain goals characterized by the reward function in the MDP while satisfying safety constraints. The typical way is to drive a potential function $V(x)$ to be zero, such as goal-reaching with $V(x)=\left\|x-x_{f}\right\|_2^2$. The RL policy will generate an action without safety guarantee first as $u_{\text{RL}}(t) = \mu\left(x_t \mid \theta^\mu\right)+\mathcal{N}_t$, where $\mu(\bullet \mid \theta^\mu)$ is a policy parameterized by deep neural networks $\theta^\mu$ and $\mathcal{N}$ is a random process for promoting exploration. \footnote{The state $x$ and $s$, the control/action $u$ and $a$, terminologies in control theory and reinforcement learning, are used interchangeably here.} Then the barrier function method \cite{ames2016control} ensures that the controller complies with the safety constraint by solving the following convex quadratic program for control synthesis 
\begin{equation}\label{eq:minnorm}
\begin{array}{ll}
\min _{u_r\in[\underline{u}, \bar{u}]} & ||u_r - u_{\text{RL}}||^2 \\
\text { s.t. } & \frac{\partial h(x)}{\partial x}(f(x)+g(x) u_r) \geq-\kappa(h(x)) \\
\end{array}
\end{equation}
where $\alpha > 0 $ and $\kappa(\bullet)$ is an extended class-$\mathcal{K}$ function (strictly increasing and $\kappa(0)=0$). Then like typical RL trajectory rollout, such process will be repeated for an episode length $T$; see Fig.~\ref{AMCBF map}(a). 
Unlike existing works in the literature using a manually designed class-$\mathcal{K}$ function, we propose to learn an extended class-$\mathcal{K}$ function parameterized by a neural network; see the illustration in Fig.~\ref{AMCBF map}(b). First, the class-$\mathcal{K}$ function is made expressive and adaptive with the parameterization of DNNs. Moreover, it should keep the property of a class-$\mathcal{K}$ function. 
(a) To make sure that $\kappa(\bullet)$ is monotonically increasing, the weights (excluding the bias) of the DNNs should be non-negative \cite{dugas2009incorporating,rashid2018qmix}, which is achieved by the absolute value (or exponential) activation function to guarantee $W_1 \geq 0$ and $W_2 \geq 0$.
(b) By setting $\kappa(z):=\kappa^{\prime}(z)-\kappa^{\prime}(0)$, we guarantee that $\kappa(0)=0$. Then the learned function $\kappa(\bullet)$ is guaranteed to be a class-$\mathcal{K}$ function.
Moreover, we consider multi-steps of CBF in the rolling-out and training process of RL policies to address the infeasibility and sub-optimality issues. The intuition is that with the learning-based multiple-step planning, 1) it can have a more global view (instead of myopic) to achieve optimality, 2) it can be more foresighted and thus avoid getting stuck into the concave trap (stability), and (3) avoid the conflicts between CBF condition and control limitations (infeasibility). Hence, with a learned class-$\mathcal{K}$ function and a \textit{multi-step training and single-step execution} paradigm, we address the three fundamental issues of CBF described in \cref{{sec:intro}} and the overall of the AM-CBF is illustrated in Fig.~\ref{AMCBF map}.

Following the RL formalism, the policy $\mu(\bullet \mid \theta^\mu)$ and the class-$\mathcal{K}$ function $\kappa(\bullet)$ will be updated to maximize the cumulative reward function as
\begin{equation}\label{loss}
J(\theta) = \sum_{k=1}^T \gamma^k {R}(s_k, a_k).
\end{equation}
Moreover, the temporal difference loss function used to train the critic network is as follows~\cite{lillicrap2015continuous} 

\begin{equation}\label{eq:critic_loss}
\begin{aligned}
&\mathcal{L}(\theta) = \mathbb{E}_{s,a,r,s^\prime}\left((y-Q(s,a|\theta^Q)\right)^2 \\
&\text{where}\ \ \ y = R + \gamma Q^{\prime}\left(s, \mu^{\prime}(s \mid \theta^{\mu^{\prime}}) \mid \theta^{Q^{\prime}}\right),
\end{aligned}
\end{equation}
where $\theta^{\mu^{\prime}}$ and $\theta^{Q^{\prime}}$ are the target networks of the actor and critic, respectively.
Gradient descent-type algorithms are used to update the parameters $\theta^\mu$, $\theta^\mathcal{K}$, and $\theta^Q$. 

\subsection{Gradient evaluation of the class-$\mathcal{K}$ function within QP via differentiable convex programming}
To update the class-$\mathcal{K}$ function, it requires to differentiate through the QP in \eqref{eq:minnorm} to get the derivative of the loss function regarding $\theta^{\mathcal{K}}$. Note that the QP in \eqref{eq:minnorm} is convex and can be differentiated via the KKT conditions \cite{amos2017optnet}, which are \textit{equivalent} conditions for (global) optimality. The KKT conditions state that at the optimal solution, the gradient of the Lagrangian function with respect to the program's input and parameters must be zero. Hence, by taking the partial derivative of the Lagrangian function with respect to the input and extending it via the chain rule to the program's parameters, we obtain all the gradients needed for training. Therefore it can be integrated seamlessly into the end-to-end training framework. 
We have integrated differentiable optimization using the cvxpylayers package \footnote{\url{https://github.com/cvxgrp/cvxpylayers}} which is an extension to the cvxpy package with an affine-solver-affine (ASA) approach. The ASA consists of taking the optimization problem's objective and constraints and mapping them to a cone program.
For a generalized QP 
\begin{equation}
\begin{aligned}
\min_x &\ \  \frac{1}{2} x^T Q x+q^T x \\
\text { s.t.} &\ \  A x=b \\
&\ \   G x \leq h,
\end{aligned}
\end{equation}
we can write the Lagrangian of the problem as:
\begin{equation}
L(z, \nu, \lambda)=\frac{1}{2} z^T Q z+q^T z+\nu^T(A z-b)+\lambda^T(G z-h)
\end{equation}
where $\nu$ are the dual variables on the equality constraints and $\lambda \geq 0$ are the dual variables on the inequality constraint.
Using the KKT conditions for stationarity, primal feasibility, and complementary slackness.
\begin{equation}
\begin{aligned}
Q z^{\star}+q+A^T \nu^{\star}+G^T \lambda^{\star} & =0 \\
A z^{\star}-b & =0 \\
D\left(\lambda^{\star}\right)\left(G z^{\star}-h\right) & =0
\end{aligned}
\end{equation}
By differentiating these conditions, we can shape the Jacobian of the problem as follows.
\begin{equation}
\left[\begin{array}{l}
d_z \\
d_\lambda \\
d_\nu
\end{array}\right]=-\left[\begin{array}{ccc}
Q & G^T D\left(\lambda^{\star}\right) & A^T \\
G & D\left(G z^{\star}-h\right) & 0 \\
A & 0 & 0
\end{array}\right]^{-1}\left[\begin{array}{c}
\left(\frac{\partial \ell}{\partial z^{\star}}\right)^T \\
0 \\
0
\end{array}\right]
\end{equation}
Furthermore, via chain rule, the derivatives of the loss function regarding any of the parameters in the QP, including the class-$\mathcal{K}$ function, are available~\cite{amos2017optnet}. This will enable end-to-end training for any learnable modules in this framework.
This differentiable programming module is integrated into DDPG training process\footnote{\url{https://github.com/philtabor/Youtube-Code-Repository/blob/master/ReinforcementLearning/PolicyGradient/DDPG/pytorch/lunar-lander/ddpg_torch.py}}. Moreover, during training, multiple tasks will be encountered and thus the resulting controller can be adaptive to different or even unseen tasks. Note that in execution, only one step of the QP in \eqref{eq:minnorm} is needed to solve (the same as normal CBF). As a result, this AM-CBF can address the critical limitations of existing CBF-based approaches and can lead to more adaptive, reliable, and safe controllers.
Algorithm~\ref{DDPGalgo} summarizes the overall framework with the DDPG~\cite{lillicrap2015continuous} and the learnable AM-CBF.

\begin{algorithm}[H]
\caption{Safe reinforcement learning with AM-CBF}\label{DDPGalgo}
\begin{algorithmic}[1]
\State \textbf{Require:} Environment setting, learning rates $\alpha, \beta$, discount factor $\gamma$, and target network update rate $\tau$
\State Initialize critic network  $Q\left(s, a \mid \theta^Q\right)$, actor $\mu\left(s \mid \theta^\mu\right)$ and $\mathcal{K}$-function network with weights $\theta^Q$ and $\theta^\mu$ and $\theta^\mathcal{K}$
\State Initialize target network $Q^{\prime}$ and $\mu^{\prime}$ with weights $\theta^{Q^{\prime}} \leftarrow \theta^Q, \theta^{\mu^{\prime}} \leftarrow \theta^\mu$ 
\State Initialize replay buffer $\mathcal{R}$
\For{episode $=1,\ldots,M$}
\State Initialize a random process $\mathcal{N}$ for action exploration
\State Receive initial observation state $s_1$ 
\For{${t}=1,\ldots,T$}
\State \parbox[t]{200pt}{Select action $a_t=\mu\left(s_t \mid \theta^\mu\right)+\mathcal{N}_t$ according to the current policy and exploration noise \strut}
\State \underline{Rectify the action via \eqref{eq:minnorm} for safe exploration}
\State \parbox[t]{200pt}{Execute action $a_{t_R}$ and observe reward $r_t$ and new state  $s_{t+1}$\strut}
\State Store transition $\left(s_t, a_t, a_{t_R}, R_t, s_{t+1}\right)$ in $\mathcal{R}$
\State \parbox[t]{200pt}{Sample a random mini-batch of ${N}$ transitions $\left(s_t, a_t, a_{t_R}, R_t, s_{t+1}\right)$ from $\mathcal{R}$\strut} 
\State \parbox[t]{200pt}{Update critic by minimizing the loss in \eqref{eq:critic_loss} with learning rate $\beta$ \strut}  
\State \parbox[t]{200pt}{Update the actor $\theta^\mu$ and $\mathcal{K}$-function $\theta^{\mathcal{K}}$ using the gradient ascent with the sampled gradient of the return in \eqref{loss} \strut}  
\State $\theta^\mu \leftarrow \theta^\mu + \alpha \nabla_{\theta^\mu}J(\theta)$
\State \underline{$\theta^{\mathcal{K}} \leftarrow \theta^{\mathcal{K}} + \alpha \nabla_{\theta^{\mathcal{K}}}J(\theta)$}
\State Update the target networks with rate $\tau$ 
\State $\theta^{\prime} \leftarrow \tau \theta+(1-\tau) \theta^\prime$ 
\EndFor
\EndFor 
\State \textbf{Return:} $\theta^\mu, \theta^{\mathcal{K}}, \theta^Q$.
\end{algorithmic}
\end{algorithm}

\section{SIMULATIONS AND RESULTS}
In this section, we evaluate the AM-CBF performance in two cases of a Dubin's car environment, a first-order and a second-order system.
Three research questions are answered, originating from the limitations of the current CBFs. We compare our approach with non-learning-based CBF with all other settings identical.

\subsection{Optimality}

To evaluate the AM-CBF performance on the optimality of trajectory, we used the first-order Dubins car environment we the following kinematics\eqref{dubinskinematics}.
\begin{equation}\label{dubinskinematics}
\left(\begin{array}{c}
\dot{x} \\
\dot{y} \\
\dot{\theta}
\end{array}\right)=\left[\begin{array}{ccc}
\cos \theta & -\sin \theta & 0 \\
\sin \theta & \cos \theta & 0 \\
0 & 0 & 1
\end{array}\right]\left(\begin{array}{c}
v_x \\
v_y \\
\omega
\end{array}\right),
\end{equation}
where $v_x$ is the velocity along the $x$ axis of the car's frame, $v_y$ is the sideways velocity, and $\omega$ is the angular velocity. In order to reach its final destination $x_f$ from an initial state $x_o$, we designed a reward that penalizes the squared distance between the car and the goal state multiplied by a coefficient as $d\left\|x-x_{f}\right\|_2^2$, and penalizes every time step by a constant $s$ for minimum time goal-reaching. Hence, the reward is defined as:
\begin{equation}
    R = -d\left\|x-x_{f}\right\|_2^2 - s,
\end{equation}
with $d>0$ and $s\ge 0$. The discount factor $\gamma$, learning rates for training the actor and critic, and the update rates for the target networks are summarized in Tables~\ref{tab:stability-constants} and \ref{tab:hyper-parameters} in the appendix.

Fig.~\ref{opttraj} presents the trajectories from both the class-$\mathcal{K}$ functions from both AM-CBF and the linear ones. It is shown that the linear CBF follows a myopic trajectory where it avoids the obstacle only after reaching it resulting in a sub-optimal path. While the AM-CBF starts the avoidance from the initial state and clears the obstacle in a more optimal way in terms of the shortest path. 

\begin{figure}[thpb]
  \centering
  \includegraphics[scale=0.5]{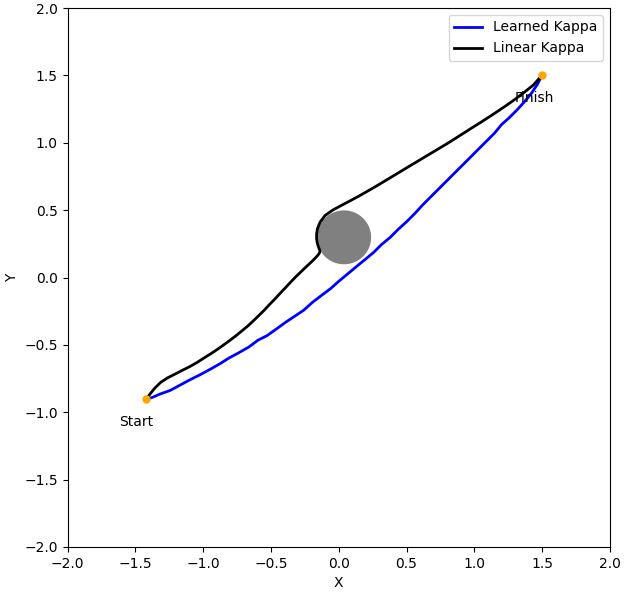}
  \caption{Dubins car trajectories for learning based AM-CBF and linear $\mathcal{K}$-function CBF.}
  \label{opttraj}
\end{figure}

\begin{figure}[thpb]
  \centering
  \includegraphics[scale=0.5]{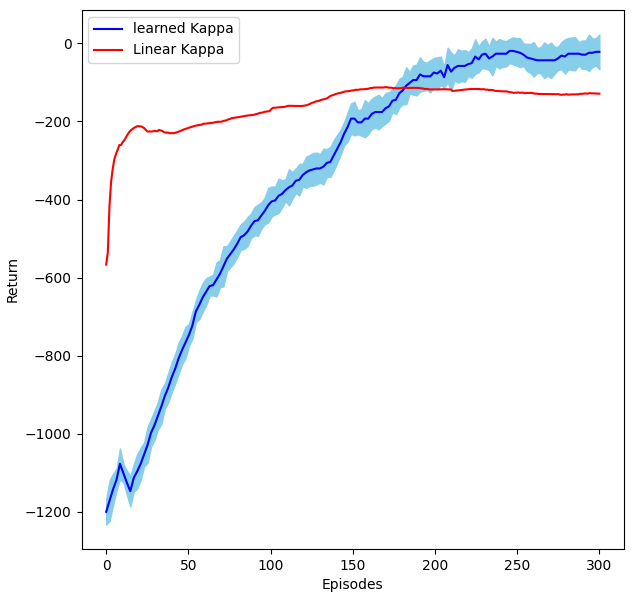}
  \caption{Return comparison for Dubins car between AM-CBF and linear $\mathcal{K}$-function CBF. The shadowed area denotes the variance from three runs with different random seeds.}
  \label{optreward}
\end{figure}

Quantitatively, the reward functions of both cases are plotted in Fig.~\ref{optreward}, where we can see that the non-learning-based CBF approach has a lower training time compared to the AM-CBF. However, the AM-CBF reaches a higher return value, which indicates the optimality of the trajectory and the shorter time to reach the final destination.

\begin{figure}[thpb]
  \centering
  \includegraphics[scale=0.5]{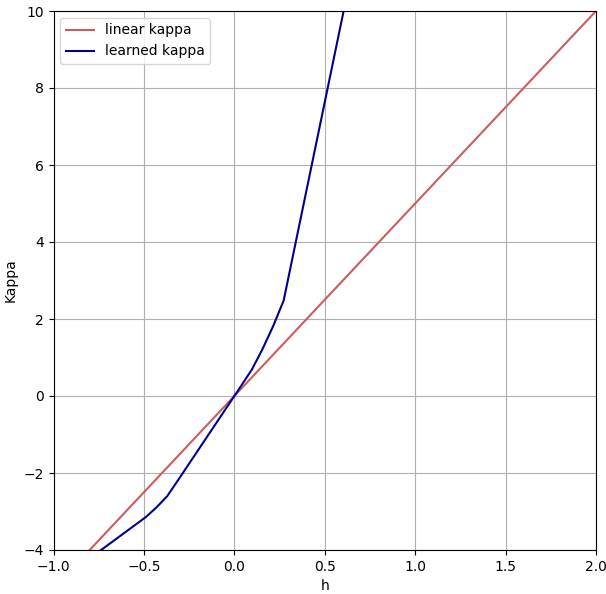}
  \caption{$\mathcal{K}$-function learned from AM-CBF and the linear $\mathcal{K}$-function.}
  \label{Kfunction}
\end{figure}

The final trained class-$\mathcal{K}$ function for the Dubins car is plotted alongside the linear function used in the normal CBF in Fig. ~\ref{Kfunction}.
Intuitively, the learned function represents a piecewise affine function in the form of an increasing quadratic function. The learned class-$\mathcal{K}$ functions share similar forms across different scenarios.

\subsection{Stability}
\begin{figure}[thpb]
  \centering
  \includegraphics[scale=0.5]{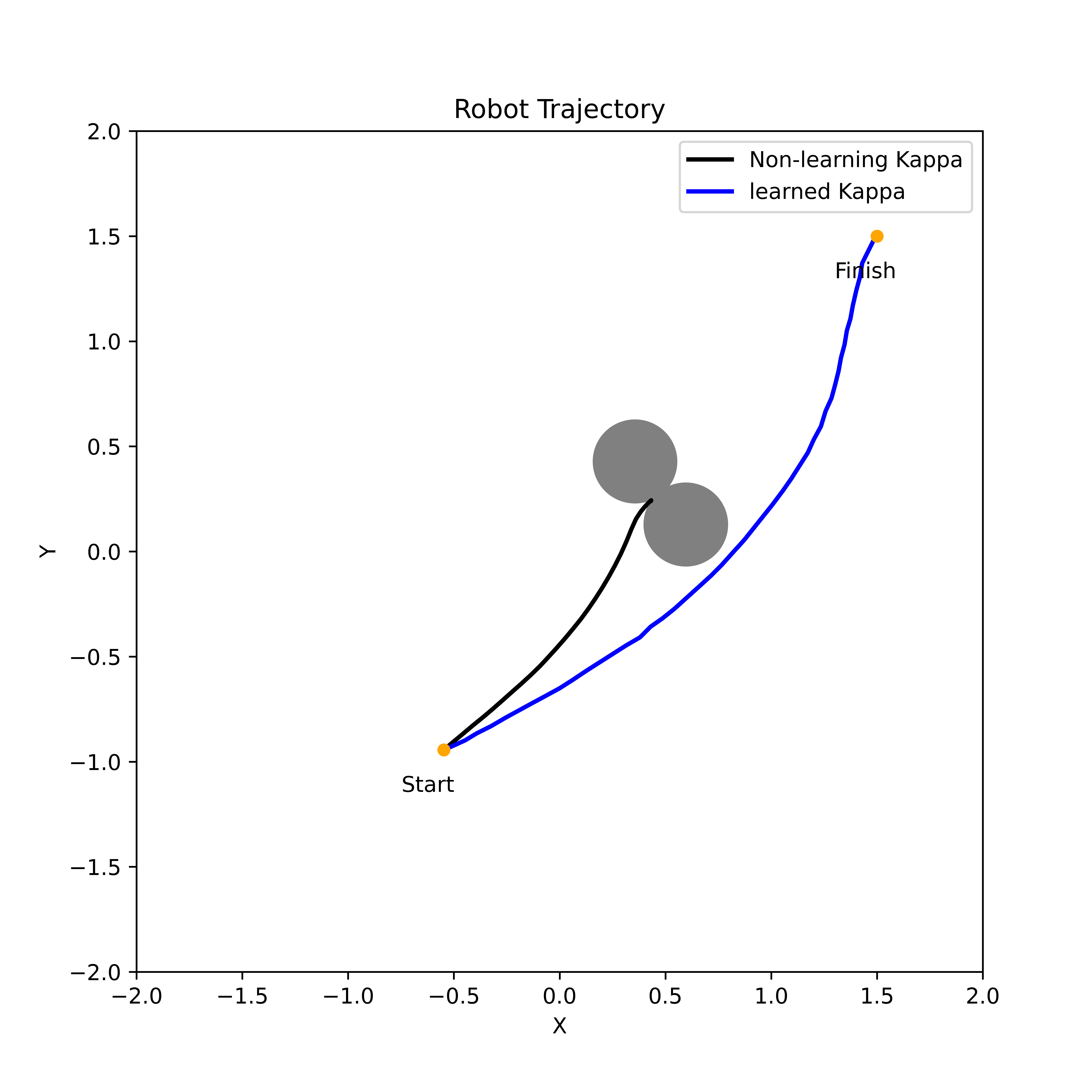}
  \caption{The AM-CBF reaches its destination, while linear $\mathcal{K}$-function getting stuck in the trap.}
  \label{stuck}
\end{figure}

To evaluate the AM-CBF performance when encountering a concave obstacle, we created two overlapped circular obstacles to create a local minimum of the Lyapunov function that can possibly trap the car. Fig.~\ref{stuck} shows how the linear CBF gets attracted to the contact point and gets stuck there, while the AM-CBF adapts and learns how to avoid the obstacle and the trap.

\begin{figure}[thpb]
  \centering
  \includegraphics[scale=0.5]{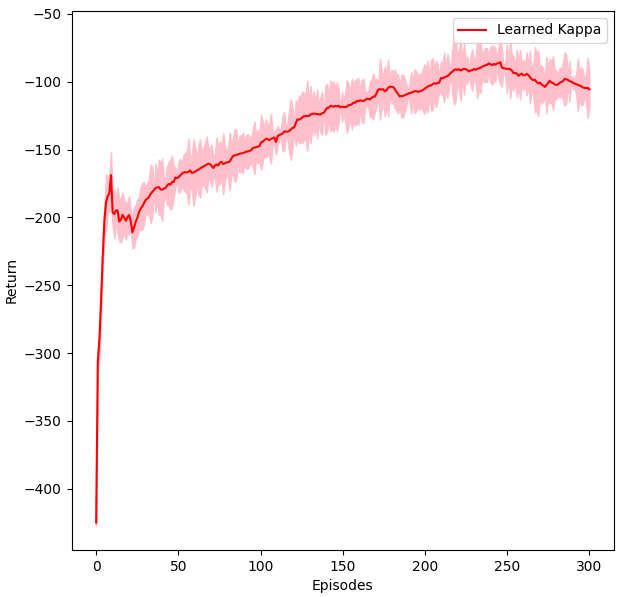}
  \caption{The return profile from AM-CBF for the stability case. The shadowed area denotes the variance from three runs with different random seeds.}
  \label{stabkreward}
\end{figure}

The return for the AM-CBF concave obstacle is plotted in Fig.~\ref{stabkreward}, we can see some instability at the beginning of the learning but it smoothes out and reaches the optimal reward.
The linear CBF has no reward profile due to the failure for reaching and thus the truncation of episodes.

\subsection{Feasibility}
\begin{figure}[thpb]
  \centering
  \includegraphics[scale=0.4]{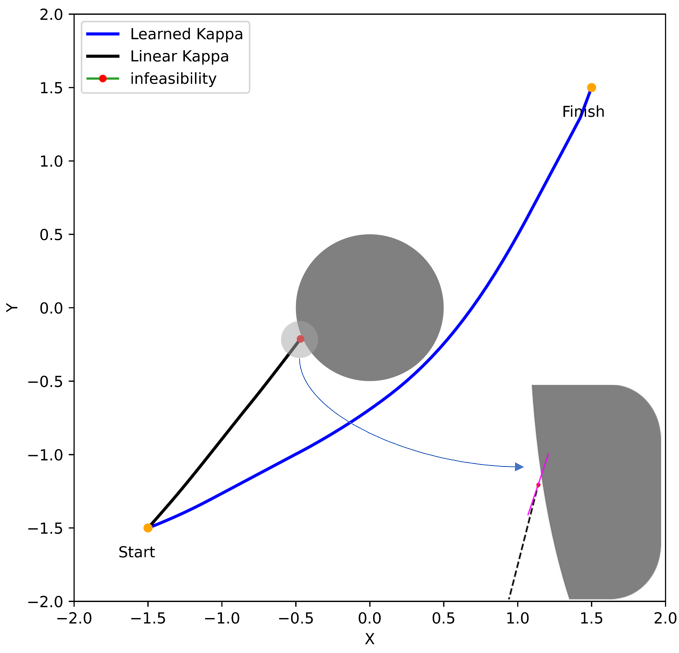}
  \caption{The AM-CBF reaching its destination in blue, and linear $\mathcal{K}$-function violates the safety constraints  next to the obstacle due to the conflicts between CBF conditions and the control constraints. The magnified magenta line shows the car's direction while encountering collision.}
  \label{inftraj}
\end{figure}

\begin{figure}[thpb]
  \centering
  \includegraphics[scale=0.5]{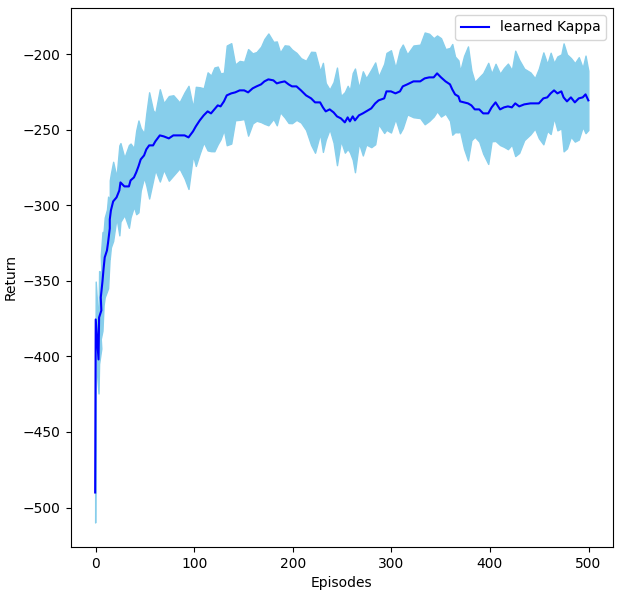}
  \caption{Return profile from AM-CBF for the feasibility case. The shadowed area denotes the variance from three runs with different random seeds.}
  \label{infreward}
\end{figure}

Infeasibility only happens in high-order systems with control constraints (e.g., upper/ lower bound). Hence, to create the infeasibility case, we use a second-order Dubin's car with the following kinematics
\begin{equation}\label{2dubinskinematics}
\left(\begin{array}{c}
\ddot{x} \\
\ddot{y} \\
\ddot{\theta}
\end{array}\right)=\left[\begin{array}{ccc}
\cos \theta & -\sin \theta & 0 \\
\sin \theta & \cos \theta & 0 \\
0 & 0 & 1
\end{array}\right]\left(\begin{array}{c}
u_x \\
u_y \\
\tau_c
\end{array}\right),
\end{equation}
with norm constraint of the control input as $\|u\| \leq u_{\max}$. We also have the following adjusted reward function to penalize the velocities at the final destination for learning to brake as well
\begin{equation}
    R = -d\left\|x-x_{f}\right\|_2^2 -b\left\|v-v_{f}\right\|_2^2 - s.
\end{equation}
In Fig.~\ref{inftraj}, it is observed that the AM-CBF avoids the obstacle by diverging earlier with constrained input, while the linear CBF only tries to avoid the obstacle after reaching it, which results in infeasibility due to constrained inputs. The side zoom-in figure shows the direction of the car when the infeasibility arises in magenta, where the translational/ rotational control inputs ($u_x, u_y, \tau_c$) are insufficient to brake/ turn enough to avoid the collision with the obstacle.
Fig.~\ref{infreward} shows the reward profile from the AM-CBF, where we can see some oscillations at the start of the learning, smoothing out as the training progresses.

\section{CONCLUSIONS}

In this paper, we proposed a novel approach to address the optimality, stability, and feasibility of control barrier functions. Our approach is called the Adaptive Multi-step Control Barrier Function (AM-CBF), where we parameterize the class-$\mathcal{K}$ function by a neural network and train it together with the reinforcement learning policy. We evaluate our method on the first and second-order Dubin's car in various scenarios, where our approach outperforms the conventional linear class-$\mathcal{K}$ function both qualitatively and quantitatively.
For future work, we plan to explore the generalization of our approach to meta-learning settings for fast adaptation to new tasks and also work on distributionally robust learning under distributional shift.




\section*{APPENDIX}
We show the hyper-parameters in learning here in Tables \ref{tab:stability-constants} and \ref{tab:hyper-parameters}.

\begin{table}[h]
\caption{The parameters used in the Dubins car}
\label{tab:stability-constants}
\begin{center}
\begin{tabular}{l|l|l}
\hline
Parameter & description & Value\\
\hline
$x_o$ & initial state & $-1.5+\text{rand},-1.5+\text{rand}, \frac{\pi}{4}$\\
\hline
$x_f$ & final state & $1.5+\text{rand},1.5+\text{rand}, \frac{\pi}{4}$\\
\hline
$d$ & distance penalty & $0.6$\\
\hline
$b$ & velocity penalty & $0.1$\\
\hline
$s$ & step penalty & $1$\\
\hline
$\gamma$ & discount factor & $0.99$\\
\hline
\end{tabular}
\end{center}
\end{table}

\begin{table}[h]
\caption{The hyper-parameters for training the neural networks}
\label{tab:hyper-parameters}
\begin{center}
\begin{tabular}{l|l}
\hline
Parameters & Value\\
\hline
Actor-Critic networks hidden layers & $(128, 64)$\\
\hline
$\mathcal{K}$-function hidden layers & $(7, 7)$\\
\hline
batch size & $64$\\
\hline
Critic learning rate ($\beta$) & $0.01$\\
\hline
Actor learning rate ($\alpha$) & $0.001$\\
\hline
Target update rate  ($\tau$) & $0.7$\\
\hline
\end{tabular}
\end{center}
\end{table}




\bibliography{root.bib}
\bibliographystyle{IEEEtran}





\balance

\end{document}